\documentclass[fleqn,10pt]{wlscirep}
\usepackage[utf8]{inputenc}
\usepackage[T1]{fontenc}
\newcommand{\citep}[1]{\cite{#1}}

\usepackage{booktabs}
\usepackage{subcaption}
\usepackage{placeins}
\usepackage{color, soul}
\renewcommand\hl[1]{#1} 
\usepackage[draft]{todonotes}   
\usepackage{lineno}

\title{Automating the assessment of biofouling in images using expert agreement as a gold standard}

\author[1,*]{Nathaniel J. Bloomfield}
\author[2]{Susan Wei}
\author[3]{Bartholomew Woodham}
\author[3]{Peter Wilkinson}
\author[1]{Andrew Robinson}
\affil[*]{nathaniel.bloomfield@unimelb.edu.au}
\affil[1]{Centre of Excellence for Biosecurity Risk Analysis, The University of Melbourne}
\affil[2]{Mathematics And Statistics, The University of Melbourne}
\affil[3]{Biosecurity Animal Division, Department of Agriculture, Water and the Environment}

\begin{abstract}
Biofouling is the accumulation of organisms on surfaces immersed in water. It is of particular concern to the international shipping industry because it increases fuel costs and presents a biosecurity risk by providing a pathway for non-indigenous marine species to establish in new areas. There is growing interest within jurisdictions to strengthen biofouling risk-management regulations, but it is expensive to conduct in-water inspections and assess the collected data to determine the biofouling state of vessel hulls. Machine learning is well suited to tackle the latter challenge, and here we apply deep learning to automate the classification of images from in-water inspections to identify the presence and severity of fouling. \hl{We combined several datasets to obtain over 10,000 images collected from in-water surveys which were annotated by a group biofouling experts. We compared the annotations from three experts on a 120-sample subset of these images, and found that they showed 89\% agreement (95\% CI: 87--92\%). Subsequent labelling of the whole dataset by one of these experts achieved similar levels of agreement with this group of experts, which we defined as performing at most 5\% worse (p=0.009--0.054). Using these expert labels, we were able to train a deep learning model that also agreed similarly with the group of experts (p=0.001--0.014), demonstrating that automated analysis of biofouling in images is feasible and effective using this method.}
\end{abstract}


\begin{document}
\flushbottom
\maketitle
\section*{Introduction}
Global trade relies on the international shipping industry, which has been implicated in the spread of many marine non-indigenous species (NIS) around the world \citep{hayes2003identifying, Clarke:2017fx}. Modern vessels have two primary pathways for translocating NIS, namely (i) as stowaways in ballast water, or (ii) attached to the vessel surface as biofouling \citep{floerl2009potential}; examples of each follow. Ballast water was the likely vector for zebra mussels to spread from Europe to the great lakes in North America \citep{johnson1996post}, where they have led to increases in toxic blue-green algae \citep{vanderploeg2001zebra} and cost industry more than \$200 million per year in maintaining water intake structures \citep{connelly2007economic}. Biofouling is one of the most significant pathways for the spread of non-indigenous seaweeds \citep{durr2009biofouling, farrell2004boats, williams2007global}, which can outcompete native species \citep{levin2002community}, make native kelp forests less resilient \citep{bulleri2017few} and adversely impact fishing and tourism operations \citep{freshwater2006distribution, meretta2012occurrence}.

Although vessels are incentivised to manage their biofouling to reduce hydrodynamic drag and fuel costs \citep{floerl2009potential, schultz2011economic}, it is a challenging undertaking and biofouling can occur even on hulls that employ current best practice\citep{davidson2010recreational, georgiades2017evidence}. The primary method of biofouling management is the regular application of anti-fouling coatings. These contain biocides, such as copper, or create a surface that releases organisms or dissuades attachment to slow down the process of biofouling accumulation \citep{chambers2006modern}. A vessel's operating profile contributes to fouling risk, with extended periods of inactivity being associated with higher biofouling pressure \citep{scardino2009fouling}. Niche areas, such as sea chests, propellers, and other complex surface structures are at high risk of becoming fouled as they can offer a sheltered environment for fouling organisms to establish. They are also a lower priority for management as they contribute less to hydrodynamic drag compared to the flat surfaces of the hull \citep{moser2017quantifying}.

There is growing interest in \hl{improving} management of the biofouling pathway by biosecurity regulators \citep{davidson2016mini, hayes2019assessment}. New Zealand has implemented a clean hull standard that sets requirements for vessels to manage biofouling and proposed a clean hull threshold to determine the potential biosecurity risk of a vessel\citep{CRMSBIOFOUL2018}. For vessels staying longer than three weeks or visiting areas other than those designated as places of first arrival, any macrofouling except for goose barnacles is considered to be a biosecurity risk, while for short stay vessels there are macrofouling coverage thresholds \citep{georgiades2017evidence, georgiades2020regulating}. In implementing this policy, they have stressed the vessel management requirements rather than the thresholds, as even with current best management practices ships can become fouled\citep{georgiades2017evidence}. Australia has also proposed requirements for vessels to implement biofouling management practices or provide evidence that their fouling is appropriately controlled \citep{hayes2019assessment}.

In-water inspections are the best way to verify biofouling standards are being met and to collect the necessary data to measure the effectiveness of biofouling management practices. However, in-water inspections are expensive, require specialist dive teams to operate in an environment with a number of health and safety risks, and while inspections are being conducted vessels are restricted in the activities they can undertake \citep{zabin2018will}. A biofouling expert also either needs to be present during the inspection, or review the images and footage gathered afterwards, which can be a costly and time-consuming process. An alternative is to employ an underwater drone or remotely operated vehicle (ROV), which would enhance data collection opportunities but also potentially increase the burden on the expert interpreting the data.

In this paper we explore the potential for deep learning, a type of machine learning which models phenomena using deep neural networks, to automate or assist the analysis of biofouling inspection data. In the last decade deep learning has revolutionised computer vision; in fact, many regard AlexNet, the 2012 winner of the ImageNet visual recognition challenge \citep{deng2009imagenet}, as the watershed moment for deep learning \citep{rawat2017deep}. AlexNet was among the first deep convolutional neural networks (CNN) \citep{krizhevsky2012imagenet}, an architecture that is particularly suited to computer vision tasks. Our present approach is motivated by the plethora of successful applications of deep CNNs to complex image recognition tasks, from identification of wild animals in camera trap images \citep{norouzzadeh2018automatically, tabak2019machine} to identification of coral species \citep{gomez2019coral}.

A prominent example of automating biofouling image analysis is CoralNet, a machine learning method initially designed for annotating benthic surveys of coral reefs using a random annotation point approach \citep{beijbom2012automated}. CoralNet has been applied to assess the level of cover of different species and higher level taxonomic groups present in fouling communities on oil platforms in the UK continental shelf, using images taken by ROVs \citep{gormley2018automated}. Our aim in this current study was to develop a method that could be used to assess biosecurity risk, and in this context CoralNet was less suitable. Most of the images of vessel hulls that were available for developing our method had limited biofouling coverage. Unlike coral reefs and oil platforms, vessels are not stationary and actively manage their biofouling. This makes sampling error an important consideration for annotation point approaches, like CoralNet, and as CNNs consider the whole image they do not have this weakness.

Determining the potential biosecurity risk of a vessel also does not require the identification of particular species. It has been found that there is a positive relationship between the degree of biofouling present on a vessel and the number of NIS present \citep{georgiades2020regulating}. This has led many jurisdictions, such as New Zealand, to require biofouling to be managed holistically rather than targeting specific species \citep{georgiades2017evidence}. Species-based approaches also scale poorly in the marine context, as there is a large number of species that can be observed in these communities, the taxonomy is highly complex, and previously unobserved species are common \citep{davidson2016mini, hayes2019assessment}. \hl{Instead, we aimed to identify the presence and severity of biofouling. This is a much simpler problem, and makes our approach more resilient to these taxonomic issues.}

\section*{Methods}
\subsection*{Dataset}

We assembled a dataset of 10,263 images collected from in-water surveys of around 300 commercial and recreational vessels. This dataset comprised images provided by three jurisdictions, namely: the Australian Department of Agriculture, Water and the Environment (DAWE), the New Zealand Ministry for Primary Industries (MPI), and the California State Lands Commission (CSLC). Examples from the CSLC dataset are available in the literature\citep{davidson2018history}, and the MPI data set has previously been used to inform vessel biofouling management in New Zealand \citep{georgiades2017evidence, bell2011risk, georgiades2020regulating}.

\hl{Each image was labelled using the six-class Level of Fouling (LoF) scheme{\citep{floerl2005risk}}, and these annotations were provided by the jurisdiction the images belonged to. Additionally, DAWE was able to provide annotations for all three datasets using a Simplified Level of Fouling (SLoF) scale (Table {\ref{tbl:slof}}). This scale was based on the LoF scheme, but collapsed the six levels into pairs to create a three-class scale. Due to inconsistencies in the LoF labelling across the three jurisdictions, we used the SLoF labels in this study. The SLoF scheme also provided the simplest possible set of annotations that supported our goal of identifying images with fouling present and highlighting images with severe fouling.}

\hl{The SLoF labels provided by DAWE consisted of two sets of annotations from several experts from Ramboll New Zealand, who held qualifications in marine biology and had extensive experience working with biofouling imagery. The first set involved three experts grading a 120 image subset of the DAWE data, constructed by stratified random sampling to ensure balance across LoF. We refer to this as the \textbf{expert-group} labels. In the second set, one of these experts graded the full amalgamated dataset of 10,263 images and we call this the \textbf{expert} labels. The examples and user interface that were used to facilitate labelling the images are given in the supporting information.} 

\begin{figure*}[ht]
  \centering
  \begin{subfigure}{0.33\textwidth}
    \centering
    \includegraphics[width=.99\linewidth]{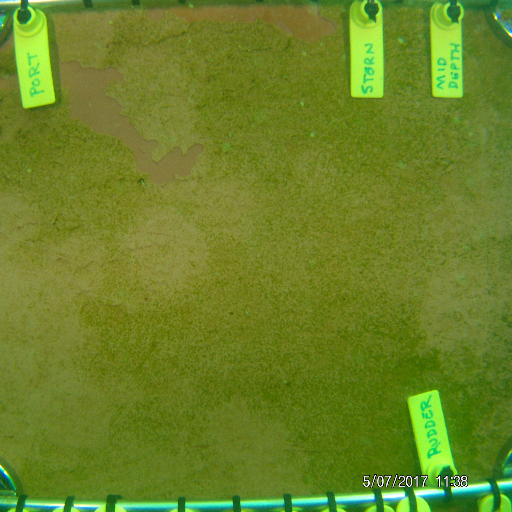}
    \includegraphics[width=.99\linewidth]{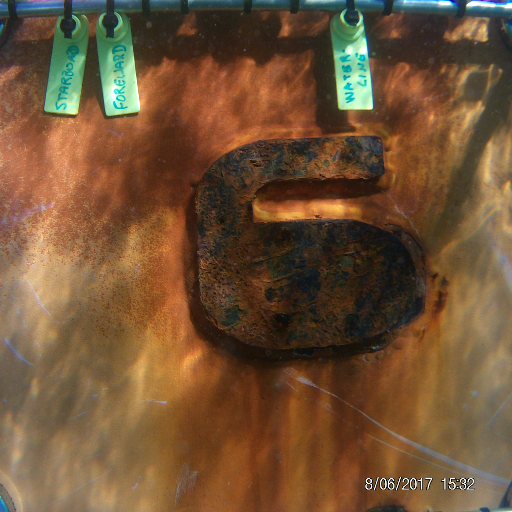}
    \caption{SLoF 0}
  \end{subfigure}
  \begin{subfigure}{0.33\textwidth}
    \centering
    \includegraphics[width=.99\linewidth]{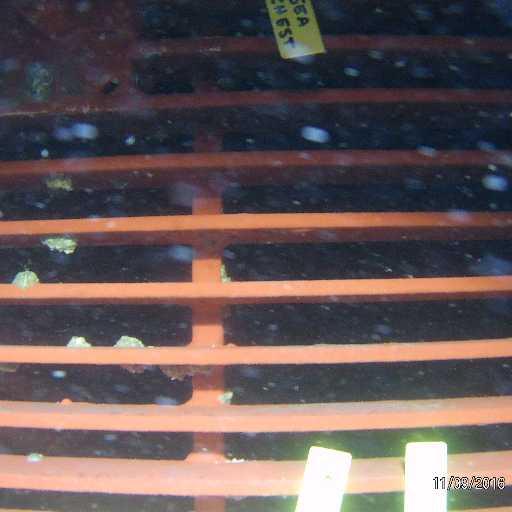}
    \includegraphics[width=.99\linewidth]{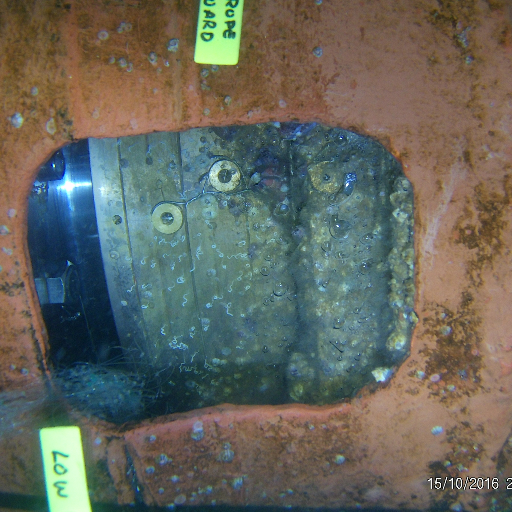}
    \caption{SLoF 1}
  \end{subfigure}%
  \begin{subfigure}{0.33\textwidth}
    \centering
    \includegraphics[width=.99\linewidth]{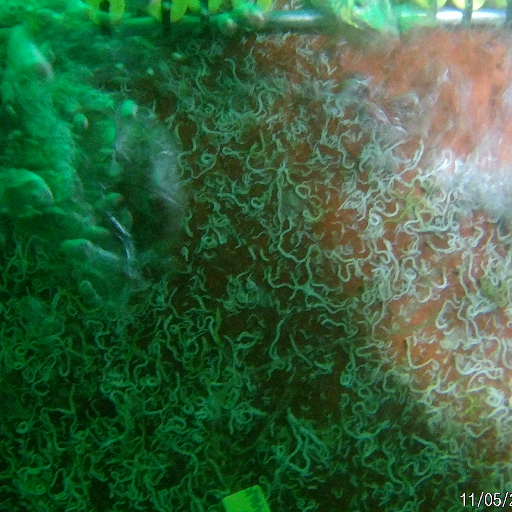}
    \includegraphics[width=.99\linewidth]{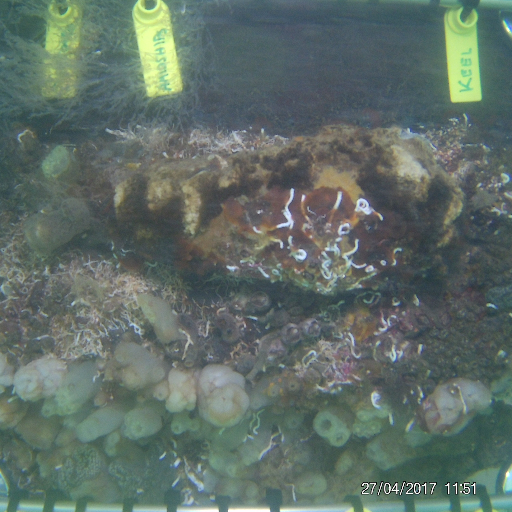}
    \caption{SLoF 2}
  \end{subfigure}
  \caption{Example images of different SLoF classes.}
  \label{fig:example_SLoF}
\end{figure*}

\hl{The SLoF labelled dataset was highly imbalanced, with most of the images being in class SLoF 0 compared to ~20\% in SLoF 1 and ~10\% in SLoF 2 (Table {\ref{tbl:traintestval}}). Example images and their SLoF labels are provided in Figure {\ref{fig:example_SLoF}}, which highlight some of the variation in the imagery in terms of lighting conditions, antifoulant coating quality, niche areas and biofouling organisms that was present in the dataset.}

\hl{We divided the overall dataset of 10,263 images into a training set and a test set, as is commonly done in machine learning to enable proper evaluation. The test set consists of the 120 expert-group images plus 721 other images from 14 vessels selected with varying degrees of fouling as determined by SLoF. The test set was constructed to challenge the machine learning model with different styles of vessel niches and fouling communities. Hence, we had a total of 841 images in the test set; the remaining data were used to both train the deep learning model and perform cross-validation (5-fold) for hyperparameter tuning.}

\subsection*{Machine Learning}
A machine learning algorithm typically learns by training on a set of examples. We present to the machine learning algorithm a set of images with the accompanying SLoF labels (i.e. 0, 1, 2). We wish the algorithm to accurately label images \textit{outside} of this training set, i.e., to generalize to never-before-seen images.

The setup so far makes the problem a classic supervised learning task. However unlike most image classification problems, our classes are ordinal. For example, mistaking an image of SLoF 2 as 0 is a larger error compared to mistaking an image of SLoF 1 as 0. This is an analogous challenge to the recent APTOS 2019 Blindness Detection Kaggle competition \citep{Kaggle2020}, which asked participants to \hl{build a model that labels the severity of a disease in images on an integer scale}. Many of the best performing Kaggle entries used regression losses rather than classification losses, and we follow the same approach here as this allows the relative magnitude of errors to be easily captured.

To measure the model performance, we consider our three-class problem as two separate binary classification tasks: 1) identify fouled images (SLoF $= 0$ versus SLoF $>0$) and 2) identify heavily fouled images (SLoF $=2$ versus SLoF $< 2$). This allows us to measure the effectiveness of our model as a classifier without choosing arbitrary class thresholds. Instead of the more commonly used receiver operating characteristic (ROC) curve, we use the average precision metric because it provides a better indication of classifier performance in the case that classes are imbalanced \citep{saito2015precision, liu2019binormal}. We apply the average precision metric to each of the two binary classification tasks, and report finally their average as an overall indicator of performance.

Given that we are working with image data, the natural deep learning architecture to use is the convolutional neural network (CNN). A CNN comprises an input layer, which in our case is an RGB image, and an output layer, which is a raw number that relates to the SLoF class of the image. Between these are multiple hidden layers, which are connected in a sequence and make up the architecture of network. Each layer performs an operation on the previous layer, such as convolutions, pooling operations, or matrix-matrix multiplications, and the nature of these operations are determined by trainable weights \citep{rawat2017deep}. The creators of AlexNet were the first to discover that stacking a large number of these layers greatly improved performance the performance of CNNs on image-recognition tasks \citep{krizhevsky2012imagenet}.

Training a CNN consists of many components including the selection of a network architecture, a method of optimising the weights of the network (optimiser), a differentiable function that describes network performance with different configurations of weights (loss function), optimiser parameters, an image augmentation pipeline and a learning rate schedule that modifies the size of each weight update over each \emph{epoch} (\emph{i.e.}, iteration through the training data). Together these components affect the quality of the trained neural network. Often the term \emph{hyperparameter} is used to refer to parameters of the optimiser, the learning scheduler, etc. The number of possible combination of these design components is incredibly large, and the available search space for determining the best combination is limited by the amount of computing power available.

We trained and tested our deep learning models with \texttt{pytorch} \citep{NEURIPS2019}, an open-source deep-learning library developed by Facebook. We began the model building process by conducting a learning rate test \citep{smith2017cyclical}, using stochastic gradient descent (SGD) as the optimiser and a default set of optimiser parameters picked from the APTOS challenge. The result of this test was used to inform a quasi-random search for the best optimiser parameters \citep{bergstra2012random}, drawing parameters from a Sobol sequence \citep{atanassov2002new} to provide more even coverage of the search space compared to random sampling. This was done by training the model for a small number of epochs, and the best sets of optimiser parameters were chosen for further exploration in addition to the default set.

We then tested performance for different combinations of the training components.  We considered mean squared error and smooth-L1 loss, which we weighted by class frequency to remove the bias introduced by the imbalance of the dataset \citep{yue2017imbalanced}. In addition to the SGD optimisation algorithm, other optimisers such as Adaptive Moment Estimation (Adam) \citep{kingma2014adam}, Rectified Adam (RAdam) \citep{liu2019variance} and Adam with a corrected weight decay algorithm (AdamW) \citep{loshchilov2017decoupled} were tested. Several learning rate schedules were examined including a multi-step learning rate decay schedule, one-cycle \citep{smith2018disciplined} and cosine annealing \citep{loshchilov2016sgdr}. In CNNs, image augmentation pipelines are important for preventing overfitting to the training data, and two different approaches with varying complexity were tried from the APTOS competition. These applied operations to our training data images that did not change their class such as rotations, random cropping, and adjusting the colour and contrast. 

We considered off-the-shelf network architectures, starting with the small \texttt{resnet18} residual network which was used to test every possible combination of the training components above. The residual network architecture was introduced to address the vanishing gradient problem in networks with large numbers of layers by allowing inputs to skip layers, and obtained first place in the 2015 ImageNet classification challenge \citep{he2016deep}. Once the best training components were identified we trained larger and more modern network architectures on larger images, allowing us to determine if increasing image size from 256$\times$256 to 512$\times$512 pixels improved performance. These architectures included the "ResNeXT" squeeze and excitation networks which built upon the residual learning idea and introduced a squeeze and excitation block that incorporates relationships between image colour channels \citep{xie2017aggregated, hu2018squeeze}. We also tested the inception architecture, which attempts to identify features at different scales in the image by applying convolution layers with several different sized kernels simultaneously \citep{szegedy2017inception}. We also considered efficient nets, which incorporate some of these previous ideas into an architecture that is designed to scale optimally and efficiently when the number of layers is increased \citep{tan2019efficientnet}. A summary of the network architectures in this paper and the \texttt{Python} packages used to implement them are provided in Table \ref{tbl:architecture}.

We used the pre-trained ImageNet weights to initialise all of our networks. \hl{These weights are created by training networks on the ImageNet database, which contains millions of images with a thousand different categories {\citep{russakovsky2015imagenet}}, and we downloaded them for each architecture through the neural network packages in Table} \ref{tbl:architecture}. This is a common practice known as transfer learning which reduces the number of epochs required to reach a performance plateau and improves results on small datasets \citep{norouzzadeh2018automatically}. All network weights were trained, except for the batch-normalisation layers, as these are best trained on large datasets like the ImageNet database. 

The final step was creating a network ensemble. This is a technique where the class of an image is predicted by multiple networks, and their outputs are combined to obtain better performance \citep{norouzzadeh2018automatically}. We took the simplest approach, which is to average the raw network output. We identified the best performing ensemble by testing the performance of every combination of network trained on a particular image size. This gave us 510 possible ensembles to test for each image resolution. The full details of the model fitting process are provided in the supporting documentation.

\subsection*{Thresholding to create a classifier}

\begin{figure*}[h]
  \centering
  \includegraphics[width=1\linewidth]{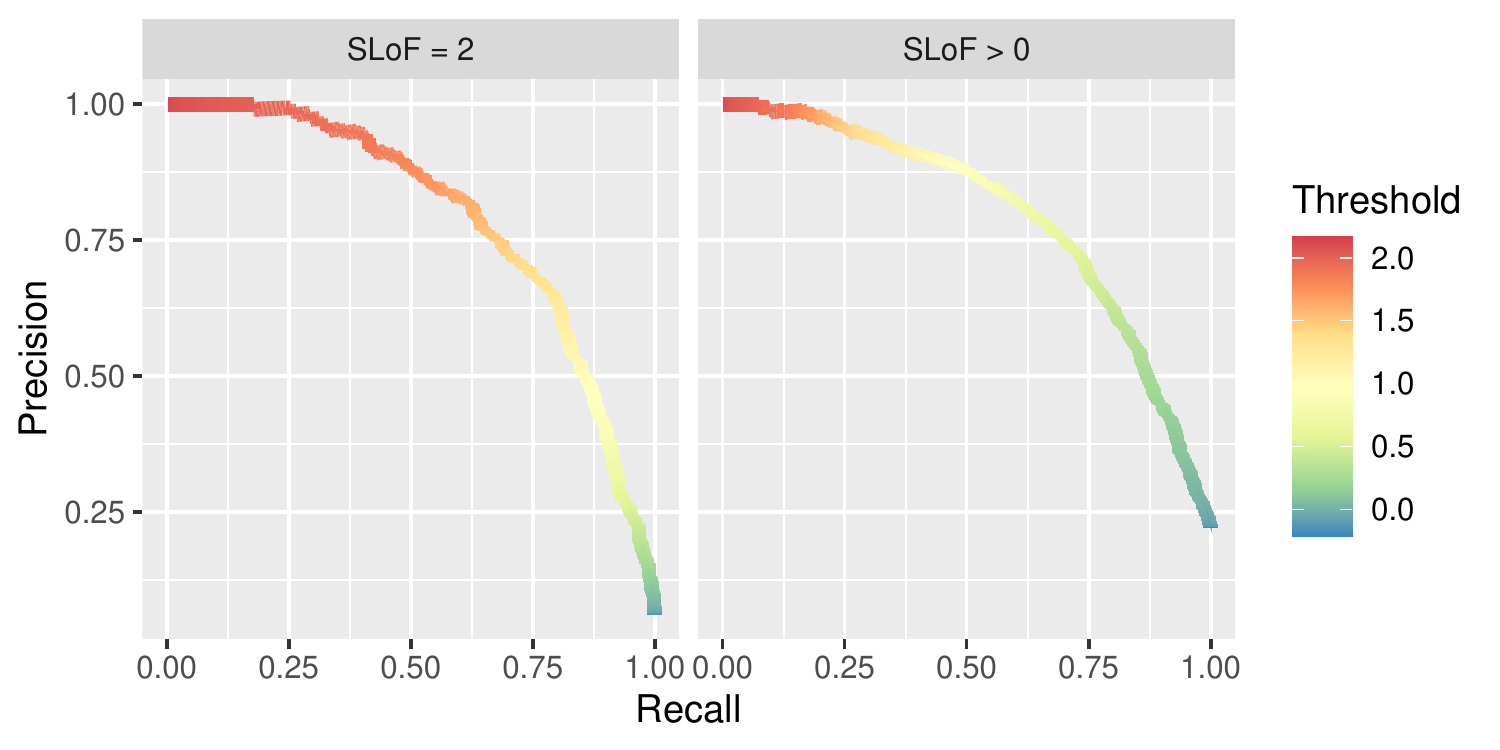}
  \caption{Precision-recall curve for model using validation data from each crossfold.}
  \label{fig:precision-recall-best}
\end{figure*}

The raw output of our model is a single number which needs to be thresholded to map back to the SLoF classes. The precision-recall curve created by combining validation crossfolds is used to guide this mapping process (Figure \ref{fig:precision-recall-best}). In particular, the curve highlights the trade-off between precision and recall when choosing a threshold. A high precision classifier will only capture some of the positive results, while a high recall classifier will capture most of the positive results along with many false positives. For illustrative purposes we have selected three classifiers to explore, \hl{namely a \textit{high-precision classifier} chosen with a 50\% recall threshold, a \textit{high-recall classifier} chosen with a 95\% recall threshold and a \textit{balanced classifier} with a 80\% recall threshold.}

\subsection*{Comparison to experts}

\hl{Perfect agreement within the SLoF labelling scheme is unlikely even among biofouling experts due to its subjectivity, and the frequency at which experts agree with each other is a useful benchmark to evaluate the performance of our models. The 120-image expert-group dataset was graded by three experts, yielding a total of 720 expert-group label pairs. These pairs were obtained by pairing the labels of one expert to annotations provided by the other two, and repeating the process for each expert. We also paired these with the expert and model labels, providing 360 label pairs to compare the performance of the expert and model to the expert-group.}

We assessed the significance of differences in precision and recall with a two-sided Fisher's exact test \citep{fisher1992statistical} with the \texttt{fisher.test} function in \texttt{R} \citep{citeR}, using the null hypothesis that the precision or recall between \hl{the expert-group labels is no different to the precision or recall with the other label sets, using the expert-group as the ground truth. We also used the two-one-sided t-tests (TOST) approach to test for non-inferiority {\citep{walker2011understanding}} using the {\texttt{TOSTER}} R package {\citep{TOSTER}}. The null hypothesis in this method was that the agreement observed in the expert-group labels would be at least 5\% better compared to the agreement observed between our other labels and the expert-group.} A separate non-inferiority test was necessary as the lack of significant differences does not mean we can conclude that two distributions are similar \citep{blackwelder1982proving}. We chose a p-value of 0.05 to signify statistical significance.

\section*{Results}

\subsection*{Model thresholding and performance}

\hl{Our best performing model based on five fold cross validation was an ensemble consisting of the} \texttt{resnet18}, \texttt{se\_resnext50\_32x4d}, \texttt{se\_resnext101\_32x4d}, \texttt{inceptionv4}, \texttt{inceptionresnetv2}, \texttt{efficientnet-b4} and \texttt{efficientnet-b5} CNN architectures (see Table \ref{tbl:architecture}) \hl{with an input image size of 512$\times$512 pixels, trained using an AdamW optimiser, tuned optimiser hyperparameters with a batch-size of 64, smooth-L1 loss, a multi-step learning rate decay schedule and the more complex set of image augmentations. This gave a final mean average precision of 0.796 (standard deviation of 0.023), which significantly improved upon the results from the hyperparameter tuning of 0.632 (standard deviation of 0.025).} The full results of the model fitting process is provided in the supporting documentation. The results for each binary classification problem with this model on our validation data and test dataset are shown in Table \ref{tbl:bestmodelres}. The classifiers show better results on the testing dataset, which is promising for the generalisability of our model.

\subsection*{Inter-rater reliability}

\begin{figure*}[ht]
  \centering
  \includegraphics[width=.99\linewidth]{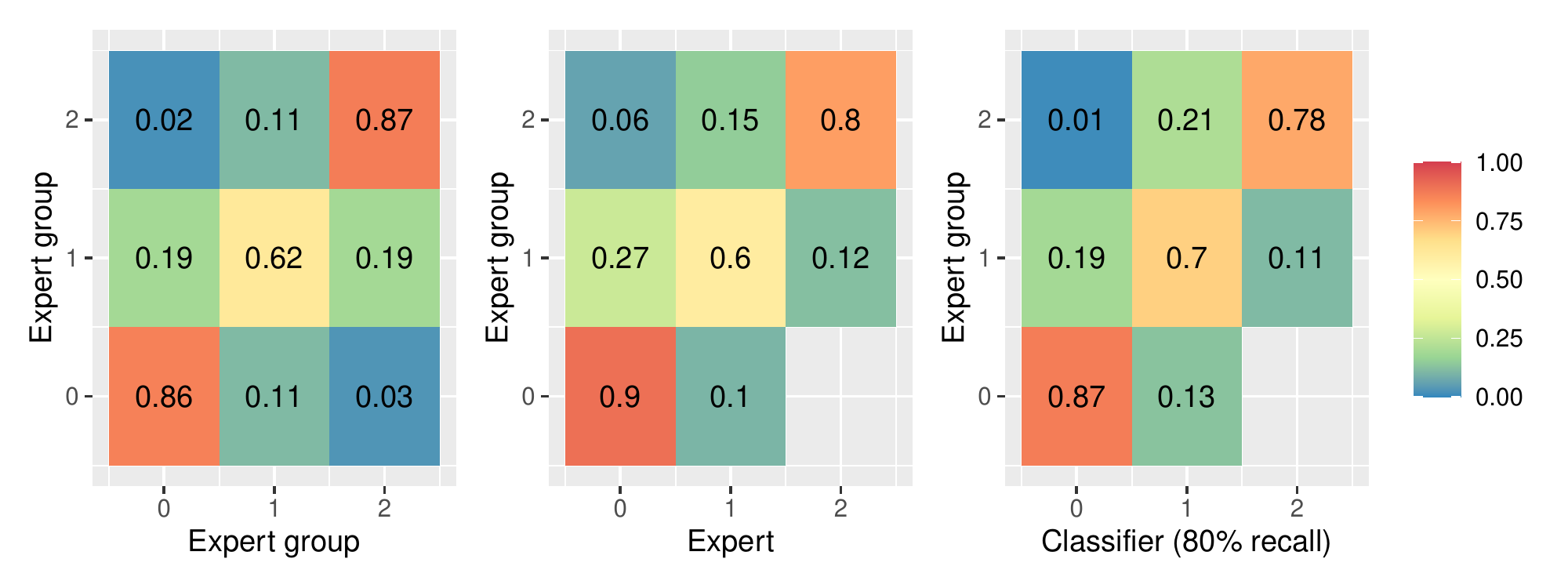}
  \caption{\hl{Confusion matrices using the SLoF score on the expert-group dataset, comparing labels between the group of experts, the expert that annnotated the full dataset, and the model annotations.}}
  \label{fig:confusion_expertvmturkvmodel}
\end{figure*}

We found that experts agree most often on images showing clean or heavily fouled hulls, while images that only contained some fouling was were more likely to obtain inconsistent grades (Figure \ref{fig:confusion_expertvmturkvmodel}). Overall, experts showed 89\% agreement for both tasks (95\% CI: 87--92\%). As we have considered every combination of experts, the recall and precision calculated for each task was the same. Experts were found to achieve 91\% precision and recall for identifying images containing fouling (95\% CI: 88--94\%) and 87\% for images containing heavy fouling (95\% CI: 82--90\%) (Table \ref{tbl:table_expertvmturk}).

\hl{When the rate of agreement between the expert and the expert-group was compared we found that the non-inferiority test showed that the agreement was similar to within a margin of at most 5\% worse (p=0.009--0.054). This similarity could also be observed from the confusion matrix between the expert-group and expert labels (Figure {\ref{fig:confusion_expertvmturkvmodel}}).}

\hl{Expert agreement is a useful benchmark for our computer vision model, and depending on the thresholds chosen to create a classifier, different outcomes were found (Table {\ref{tbl:table_expertvmturk}}). Choosing an 80\% recall threshold for both tasks resulted in a classifier with similar agreement to experts to within a margin of at most 5\% worse (p = 0.001--0.0014). The results for this classifier are shown in Figure {\ref{fig:confusion_expertvmturkvmodel}} as a confusion matrix. Using a 95\% recall threshold instead could produce significantly higher recall with respect to the expert-group labels (p=$<0.001$--0.004) at the cost of significantly lower precision (p$<0.001$). Conversely, using a much lower recall threshold of 50\% results in significantly higher precision (p=0.001--0.036), with a corresponding decrease in recall (p$<0.001$).}

\section*{Discussion}

\hl{In this study we applied deep learning methods to identify the presence and severity of biofouling on ship hulls, and compared our performance to a group of experts. We were able to train neural networks that obtain similar results to these experts, which is highly promising as it suggests that under the study conditions, automated analysis of biofouling in images is feasible and effective.} We have also demonstrated that if high precision or recall is desired for the application of the model, then classifiers can be created that offer better performance than experts with regard to this property. This allows the behaviour of the classifiers to be tuned for a particular application. For example, when screening vessels for biosecurity risk it may be desirable to have a classifier with higher recall so few images with severe fouling are missed. Conversely, if an activity were being undertaken where intervention capacity was limited then a classifier with higher precision would be more appropriate. 

\hl{In practice, the performance of the method will be impacted by image quality, lighting and water turbidity. The dataset that we used to train and test the model is diverse and included images taken under a range of different conditions as shown in Figure} \ref{fig:example_SLoF} \hl{and the supporting information. It is encouraging that despite this the model was able to obtain close to expert accuracy, suggesting that our approach can account for fouling in poor quality images within those labelled by the expert-group about as well as an expert can. This may be a best-case scenario, and performance will be different under other operational settings. The deep convolutional neural networks that we have trained can easily be fine-tuned to incorporate more challenging examples if needed, and the models can be readily adapted to the required conditions.}

\hl{Fine-tuning with local examples will also likely improve performance when deploying the model in areas where no training data was collected, as the fouling communities present on vessels may be different to those in the training dataset. However, this issue is somewhat mitigated by our focus on classifying the overall coverage of biofouling in an image, rather than just the species present. This also allows us to side-step the challenge of identifying species or species groups, which would have likely required a larger set of images to obtain similar results}\cite{gormley2018automated}\hl{.  Our approach of simply looking at the overall level of fouling is more robust for the purpose of identifying biosecurity risk, as even if a particular type of organism occurs infrequently or not at all in the training data, our models may generalise information gained from other types of biofouling to detect that the hull is still fouled.}

The effectiveness of management activities for vessel biofouling in reducing biosecurity risk is currently a key knowledge gap for regulators, which makes it difficult to determine which combination of activities will provide confidence that a vessel is low risk. This model could be applied to provide a cheaper and more reliable way to identify the most effective management strategies, if combined with standardised vessel sampling protocols\citep{zabin2018will, Georgiades2020Conduct}, clear definitions of vessel biosecurity risk, such as the clean hull standard for New Zealand\citep{georgiades2017evidence}, collection of management data, and ongoing in-water vessel inspections. This will also support more consistent assessment of effective management strategies between different organisations, which is a limitation of expert assessments. Building this evidence base would also provide benefits to industry, as it would be a basis from which to work towards regulatory alignment between different jurisdictions.

In-water cleaning and hull grooming are increasingly important biofouling management activities, as regular cleaning can limit biofouling accumulation and provide options where the anti-fouling coatings of vessels are no longer effective or have failed \citep{hunsucker2019specialized, tribou2017effects}. However, it also presents a biosecurity risk because cleaning can lead to the release of viable propagules and organisms can detach and still be viable \citep{hopkins2008management, tamburri2020water, scianni2019vessel}. One way this risk can be managed is by considering the biofouling state of the vessel before setting conditions on in-water cleaning or grooming activities. For example, New Zealand recommends that in-water cleaning of macrofouling with an international origin must capture biological waste and dispose of it on land or be rendered non-viable, but this would not be necessary if only a slime layer were present \citep{georgiades2018technical}. Automatic detection of biofouling using the state-of-the-art deep learning tools developed in this paper could be a cost-effective and reliable way for regulators and industry to process the outcomes of biofouling inspections for this purpose.

So far we have only tested our model on static images. Since videos are constructed using a stream of images, our model should be readily adaptable to videos as well. However, further work is needed to address issues such as identifying the frames in which the camera is directed towards a vessel hull as opposed to open water or where image quality is poor, which is a common issue when analysing stills obtained from ROV footage \citep{gormley2018automated}. The video format would also offer the opportunity to incorporate information from future and previous frames to improve and smooth fouling estimates, and ideas from current action recognition methods could potentially be applied \citep{shi2017sequential}. 

Our SLoF labelling scheme also only relates to the percentage cover of macrofouling present within an image, which could be more rigorously used to determine the absolute biosecurity risk of a vessel if the area of hull captured within the image could be estimated. Given that in-water inspection methods are expected to vary greatly between jurisdictions, being able to do this without the presence of scale bars would be a major advantage. One possibility would be training a deep neural network on images of vessel hulls taken using multiple cameras, and building a model that will estimate depth given a single image \citep{mal2018sparse}.

\bibliography{mybibfile}

\section*{Acknowledgements}

We would like to thank Jason Garwood and Steven Lane for their work developing the initial project case. We would like to thank Serena Orr, Dan McClary and Emily Jones from Ramboll New Zealand for classifying the image dataset. We would like to thank Anca Hanea for providing advice in comparing different labelling schemes, and Edith Arndt for insightful discussions on the biofouling literature. We would also like to thank Dan Kluza from the New Zealand Ministry for Primary Industries and Chris Scianni from the California State Lands Commission for providing their datasets. This research was undertaken using the LIEF HPC-GPGPU Facility hosted at the University of Melbourne. This Facility was established with the assistance of LIEF Grant LE170100200. This study was undertaken with the assistance of resources and services from the National Computational Infrastructure (NCI), which is supported by the Australian Government. NB and AR's contributions were also funded by the Centre of Excellence for Biosecurity Risk Analysis at the University of Melbourne.
\section*{Data Availability}

The data that support the findings of this study are available from DAWE, MPI and CSLC, but restrictions apply to the availability of these data, which were used under license for the current study, and so are not publicly available. Data are however available from the authors upon reasonable request and with permission of the respective organisations.

\section*{Author contributions statement}

NB wrote draft manuscript, curated data and coded and performed all analyses. SW advised on machine learning and access to supercomputer resources. AR advised on statistical analysis and project direction. BW and PW solicited data and advised on project direction. All authors reviewed the manuscript.

\section*{Additional Information}

The authors declare no competing interests.

\FloatBarrier
\clearpage

\begin{table*}[h]
\caption{Simplified Level of Fouling (SLoF) scale}
\label{tbl:slof}
\begin{tabular}{l|p{13cm}}
\toprule
Class & Description \\
\midrule
0 & No fouling organisms, but biofilm or slime may be present.  \\
1 & Fouling organisms (e.g. barnacles, mussels, seaweed or tubeworms) are visible but patchy (1-15\% of surface covered).  \\
2 & A large number of fouling organisms are present (16-100\% of surface covered).  \\
\bottomrule
\end{tabular}
\end{table*}

\begin{table*}[h]
\caption{Breakdown of the number of images by SLoF in the crossvalidation and test datasets.}
\centering
\label{tbl:traintestval}
\begin{tabular}{rrrr}
\toprule
 & Crossvalidation dataset & Test dataset & Total \\ 
\midrule
Images with SLoF 0 & 7328 & 494 & 7822 \\ 
  Images with SLoF 1 & 1503 & 193 & 1696 \\ 
  Images with SLoF 2 & 591 & 154 & 745 \\ 
  Total & 9422 & 841 & 10263 \\ 
\bottomrule
\end{tabular}
\end{table*}

\begin{table*}[h]
\caption{Summary of neural network architectures used in model building.}
\label{tbl:architecture}
\footnotesize
\begin{tabular}{llllll}
\toprule
Network Family & Network Architecture & Source package  & Reference & Layers & Trainable weights ($10^6$) \\
\midrule
Residual learning & \texttt{resnet18} & \texttt{torchvision} & \citep{he2016deep} & 18 & 11 \\
Squeeze and excitation &\texttt{se\_resnext50\_32x4d} & \texttt{pretrainedmodels} & \citep{xie2017aggregated, hu2018squeeze} & 50 & 25 \\
Squeeze and excitation &\texttt{se\_resnext101\_32x4d} & \texttt{pretrainedmodels} & \citep{xie2017aggregated, hu2018squeeze} & 101 & 47 \\
Inception &\texttt{inceptionv4} & \texttt{pretrainedmodels} & \citep{szegedy2017inception} & 150 & 41 \\
Inception &\texttt{inceptionresnetv2} & \texttt{pretrainedmodels} & \citep{szegedy2017inception} & 245 & 54 \\
Efficient-net &\texttt{efficientnet-b3} & \texttt{efficientnet-pytorch} & \citep{tan2019efficientnet} & 27 & 11 \\
Efficient-net &\texttt{efficientnet-b4} & \texttt{efficientnet-pytorch} & \citep{tan2019efficientnet} & 33 & 17 \\
Efficient-net &\texttt{efficientnet-b5} & \texttt{efficientnet-pytorch} & \citep{tan2019efficientnet} & 39 & 28 \\
\bottomrule
\end{tabular}
\end{table*}

\begin{table*}[h]
\caption{Precision and recall of classifier using model with chosen recall thresholds on the validation and testing dataset.}
\centering
\label{tbl:bestmodelres}
\begin{tabular}{llrrr}
\toprule
Data & SLoF & Threshold & Precision & Recall \\ 
\midrule
Validation & $> 0$ & 0.196 & 0.448 & 0.900 \\ 
  Test & $> 0$ & 0.196 & 0.638 & 0.960 \\ 
  Validation & $> 0$ & 0.413 & 0.626 & 0.800 \\ 
  Test & $> 0$ & 0.413 & 0.754 & 0.882 \\ 
  Validation & $> 0$ & 0.943 & 0.881 & 0.500 \\ 
  Test & $> 0$ & 0.943 & 0.976 & 0.588 \\ 
\midrule
  Validation & $= 2$ & 0.889 & 0.408 & 0.900 \\ 
  Test & $= 2$ & 0.889 & 0.643 & 0.935 \\ 
  Validation & $= 2$ & 1.242 & 0.632 & 0.800 \\ 
  Test & $= 2$ & 1.242 & 0.884 & 0.844 \\ 
  Validation & $= 2$ & 1.786 & 0.881 & 0.499 \\ 
  Test & $= 2$ & 1.786 & 0.946 & 0.455 \\ 
\bottomrule
\end{tabular}
\end{table*}


\begin{table*}[h]
  \caption{\hl{Precision and recall for expert-group, expert versus expert-group and classifier versus expert-group label pairs. Numbers in brackets are the 95\% confidence intervals. The TOST column are non-inferiority testing p-values using the two-one-sided t-tests approach, with the null hypothesis being that the agreement observed between experts would be at least 5\% better compared to the agreement observed for the method-expert label pairs. The p-value columns are given by a two-sided exact Fisher test, with the null hypothesis being that the method versus expert-group label pairs do not differ in their precision or recall compared to the expert-group label pairs.}}
\centering
\label{tbl:table_expertvmturk}
\small
\begin{tabular}{lllllllll}
\toprule
Labels & Classification & Label pairs & Agreement & TOST & Recall & p-value & Precision & p-value \\ 
\midrule
Expert group & SLoF $> 0$ & 720 & 0.89 (0.87-0.92) &  & 0.91 (0.88-0.94) &  & 0.91 (0.88-0.94) &  \\ 
  Expert & SLoF $> 0$ & 360 & 0.88 (0.84-0.91) & 0.054 & 0.87 (0.81-0.91) & 0.057 & 0.93 (0.89-0.96) & 0.535 \\ 
  95\% Recall Classifier & SLoF $> 0$ & 360 & 0.78 (0.73-0.82) & 0.995 & 1.00 (0.98-1.00) & $<0.001$ & 0.74 (0.69-0.79) & $<0.001$ \\ 
  80\% Recall Classifier & SLoF $> 0$ & 360 & 0.91 (0.87-0.93) & 0.001 & 0.93 (0.89-0.96) & 0.652 & 0.92 (0.88-0.95) & 0.883 \\ 
  50\% Recall Classifier & SLoF $> 0$ & 360 & 0.81 (0.77-0.85) & 0.921 & 0.70 (0.64-0.76) & $<0.001$ & 0.99 (0.96-1.00) & 0.001 \\ 
\midrule
  Expert group & SLoF $= 2$ & 720 & 0.89 (0.87-0.92) &  & 0.87 (0.82-0.90) &  & 0.87 (0.82-0.90) &  \\ 
  Expert & SLoF $= 2$ & 360 & 0.89 (0.85-0.92) & 0.009 & 0.80 (0.72-0.86) & 0.067 & 0.92 (0.86-0.96) & 0.180 \\ 
  95\% Recall Classifier & SLoF $= 2$ & 360 & 0.78 (0.73-0.82) & 0.996 & 0.96 (0.91-0.98) & 0.004 & 0.65 (0.58-0.71) & $<0.001$ \\ 
  80\% Recall Classifier & SLoF $= 2$ & 360 & 0.89 (0.85-0.92) & 0.014 & 0.78 (0.70-0.85) & 0.036 & 0.93 (0.86-0.97) & 0.125 \\ 
  50\% Recall Classifier & SLoF $= 2$ & 360 & 0.78 (0.73-0.82) & 0.995 & 0.46 (0.38-0.55) & $<0.001$ & 0.96 (0.88-0.99) & 0.036 \\ 
\bottomrule
\end{tabular}
\end{table*}

\end{document}